\documentclass[letterpaper, 10 pt, conference]{ieeeconf}
\IEEEoverridecommandlockouts
\usepackage{booktabs}
\usepackage{multirow}
\usepackage{pifont}
\usepackage{graphicx}
\usepackage{hyperref}

\newcommand{\cmark}{\ding{51}}
\newcommand{\xmark}{\ding{55}}

\title{\LARGE \bf
Technical Report for ICRA 2026 GOOSE 2D Fine-Grained Semantic Segmentation Challenge: Exploring Query-Based Segmentation and Increased Spatial Context for Outdoor Scene Understanding
}

\author{
David Pascual-Hernández$^{1,*}$, Roberto Calvo-Palomino$^{1}$, Inmaculada Mora-Jiménez$^{1}$, Jose María Cañas-Plaza$^{1}$%
\thanks{*This work was supported by Spanish Agencies AEI and CDTI as part of the Ministry of Science, Innovation and Universities with grant number PLEC2023-010303 (GAIA); by Comunidad de Madrid with grant number TEC-2024/TEC-62 (iROBOCITY 2030-CM); and by the Spanish Agency AEI with Grant number PID2022-136887NB-I00 (POLIGRAPH).}%
\thanks{$^{1}$Authors are with Department of Signal Theory and Communications, Telematics and Computing Systems, Rey Juan Carlos University, Fuenlabrada, 28943, Spain.}%
\thanks{$^{*}$ Corresponding author: \tt\small david.pascualhe@urjc.es}%
}

\begin{document}

\maketitle
\thispagestyle{empty}
\pagestyle{empty}

\begin{abstract}

In this report, we present our submission to the GOOSE 2D Fine-Grained Semantic Segmentation Challenge, organized as part of the Workshop on Field Robotics at ICRA 2026. The challenge combines data from the GOOSE and GOOSE-Ex datasets, which comprise more than 13k images captured from 4 distinct camera setups, annotated using a hierarchical taxonomy of 56 fine-grained classes and 11 broader categories. Starting from SegFormer as a baseline, we progressively improve segmentation performance through increased training crop sizes, a transition to the query-based Mask2Former architecture, and test-time augmentation. Our experiments show that query-based segmentation significantly outperforms the baseline model. Furthermore, increasing the crop size used during training yields substantial gains, highlighting the relevance of preserving scene context for fine-grained semantic disambiguation. Our final submission, using test-time augmentation, achieves an mIoU of 69.6\% on the challenge test set, providing a strong baseline for fine-grained semantic segmentation in outdoor environments. To facilitate reproducibility and future research, code and weights will be made publicly available at \url{https://github.com/RoboticsLabURJC/outdoor-fine-grained-segmentation}.

\end{abstract}

\section{INTRODUCTION}

Fine-grained semantic segmentation plays a key role in safe navigation in unstructured outdoor environments, where autonomous vehicles require very detailed scene parsing capabilities to accurately assess terrain traversability and potential hazards. Compared with urban settings, outdoor scenes present unique challenges, such as diffuse boundaries between elements, ambiguous categorizations, and high variability in terrain types and obstacles~\cite{min2026autonomous}. Furthermore, the limited availability of annotated data has hindered research progress in addressing the challenges posed by outdoor scene understanding.

To bridge this gap, the GOOSE~\cite{mortimer2024goose} and GOOSE-Ex~\cite{hagmanns2025excavating} datasets provide a unified semantic taxonomy and more than 13k densely annotated images, encompassing a wide range of environments, robotic platforms, capture setups, and sensing modalities. Organized as part of the Workshop on Field Robotics in ICRA 2026, the GOOSE 2D Fine-Grained Semantic Segmentation Challenge builds upon these datasets and prompts participants to address several fundamental challenges of outdoor scene understanding, namely:
\begin{itemize}
    \item \textbf{Fine-grained semantic understanding}, which requires discrimination between a large number of visually similar terrain and obstacle classes.
    \item \textbf{Long-tailed class distributions}, where a small number of classes are widely present in the dataset while many categories are severely underrepresented.
    \item \textbf{Cross-domain robustness}, requiring strong generalization capabilities across environments, weather conditions, and robotic platforms.
\end{itemize}

In order to address these challenges, we establish a strong query-based segmentation baseline using Mask2Former~\cite{cheng2022masked} and investigate the role of spatial context during training for fine-grained outdoor semantic segmentation. More specifically, we (1) compare Mask2Former against SegFormer~\cite{xie2021segformer}, demonstrating the effectiveness of query-based mask classification approaches compared to conventional dense prediction architectures for fine-grained scene understanding; (2) perform an empirical analysis of the impact of increasing crop size during training as a means of preserving spatial context; and (3) present our final submission, which includes test-time augmentation (TTA), and achieved a 5th place ranking in the final stage of the competition.

\section{METHOD}

\subsection{Baseline: SegFormer}
Previous studies have demonstrated the superiority of Transformer-based models compared to traditional convolutional approaches for semantic segmentation in outdoor environments~\cite{pascual2026cross}. Motivated by these findings, we leverage the widely used SegFormer as a strong and computationally efficient baseline for enabling rapid exploration of different training configurations. Its architecture consists of a hierarchical Transformer encoder that extracts multi-scale features, followed by a lightweight MLP decoder that aggregates said features and produces dense semantic predictions. In our experiments, we use the B2-sized SegFormer variant, pretrained on the Cityscapes dataset~\cite{cordts2016cityscapes}, using the publicly available weights and implementation from the HuggingFace Transformers library\footnote{\url{https://huggingface.co/nvidia/segformer-b2-finetuned-cityscapes-1024-1024}}.

\subsection{Mask2Former}
Building also on Transformer-based backbones, Mask2Former performs segmentation as a query-based mask classification problem. Its architecture combines a pixel decoder that preserves high-resolution spatial information with a Transformer decoder that learns a fixed set of queries. Through masked attention, each query focuses on relevant regions in the image and predicts a segmentation mask along with its corresponding semantic category, enabling universal semantic segmentation. Given its design, we hypothesize that Mask2Former is particularly well suited for fine-grained outdoor segmentation. Instead of performing pixel-level classification, it reasons over coherent semantic regions, leveraging object extent, shape, and context to better disambiguate visually similar classes while preserving high-resolution details. In our experiments, we use the publicly available weights and implementation from the mmsegmentation library\footnote{\url{https://github.com/open-mmlab/mmsegmentation/blob/main/configs/mask2former}}, pretrained on the Cityscapes dataset with a Swin-L~\cite{liu2021swin} backbone.

\subsection{Training Configuration}\label{subsec:config}
During training, models are optimized using AdamW~\cite{loshchilov2019decoupled} with a learning rate schedule consisting of a 5-epoch warm-up until reaching the maximum learning rate of $10^{-4}$, followed by a cosine decay until the end of training at 50 epochs. As defined in the original implementations, SegFormer is trained using a pixel-wise cross-entropy loss, while Mask2Former uses a weighted combination of cross-entropy classification loss, binary cross-entropy mask loss, and Dice loss.

To enhance model generalization capabilities, we design a data augmentation pipeline consisting of random rescaling with a scale factor uniformly sampled from $[0.75, 1.25]$, random horizontal flipping with probability $p=0.5$, color jittering for brightness, contrast, and saturation in the range $[0.8, 1.2]$, and hue jittering in the range $[-0.075, 0.075]$ with probability $p=0.5$. Additionally, with probability $p=0.5$, one or more image quality perturbations are applied: Gaussian noise with $\mu=0$ and $\sigma \in [0.01, 0.05]$, Gaussian blur with kernel size 3 and $\sigma \in [0.1, 1.0]$, and random sharpness adjustment with a factor sampled from $[1, 2]$.

Furthermore, training samples are cropped to a fixed resolution of either $512\times512$ or $1024\times1024$ pixels and padded as needed. Increasing the crop size allows the model to attend to regions further away during training, enabling access to scene-level contextual cues that help disambiguate visually similar semantic categories.

\section{EXPERIMENTS}

\subsection{Experimental Setup and Protocol}
All experiments are conducted on an NVIDIA A100 GPU and training is performed with a batch size of 16. In the case of training with larger crops, the batch size is reduced to fit GPU constraints and gradient accumulation is used to keep an effective batch size of 16, ensuring fair comparison across configurations. Model training is performed using the Trainer API from the HuggingFace Transformers library.

For model benchmarking prior to submission, we use the validation split as defined by the challenge organizers, which comprises 1369 annotated images. Performance is evaluated using mean Intersection over Union ($mIoU$). Following the challenge protocol, the primary metric used in our analysis is the composite $mIoU$ or $mIoU_{comp}$, defined as the average of $mIoU_{coarse}$ and $mIoU_{fine}$. In that way, performance is evaluated at two granularity levels.

\subsection{Quantitative Analysis}

In Table~\ref{tab:ablation}, a summary of our progression from the initial SegFormer baseline to our final Mask2Former submission is shown, including the impact of increasing training crop size and TTA. All benchmarked configurations have been trained using the configuration described in Section~\ref{subsec:config} and evaluated on the validation split. The first observation is the significant improvement achieved by simply increasing the training crop size from $512\times512$ to $1024\times1024$ pixels. For SegFormer, the $mIoU_{comp}$ jumps from 58.4\% to 61.5\% (+3.1), which highlights the relevance of preserving as much spatial context as possible during training. In this way, the model is able to exploit broader scene-level cues, which can be necessary to separate visually similar elements that belong to different semantic categories.

\begin{table}[t]
    \centering
    \caption{Ablation study of our submission, from the SegFormer baseline to Mask2Former, with and without large-crop training (LCT) and test-time augmentation (TTA) for the final submission.}
    \label{tab:ablation}
    \begin{tabular}{l ccc ccc}
        \toprule
        
        \multirow{2}{*}{Model} &
        \multicolumn{2}{c}{Configuration} &
        \multicolumn{3}{c}{mIoU (\%)} \\
        
        \cmidrule(lr){2-3}
        \cmidrule(lr){4-6}
        & LCT & TTA & Comp. & Fine & Coarse \\
        
        \midrule
        
        SegFormer            & \xmark & \xmark & 58.4 & 50.8 & 66.0 \\
        SegFormer            & \cmark & \xmark & 61.5 & 54.2 & 68.7 \\
        Mask2Former          & \xmark & \xmark & 67.5 & 59.5 & 75.4 \\
        Mask2Former          & \cmark & \xmark & 73.4 & 66.9 & 79.9 \\
        \textbf{Mask2Former} & \cmark & \cmark & \textbf{74.8} & \textbf{67.9} & \textbf{81.7} \\
        
        \bottomrule
    \end{tabular}
\end{table}

An even larger improvement is achieved when transitioning from SegFormer, a dense pixel-wise classification architecture, to Mask2Former, a query-based mask classification approach. For the $512\times512$ crop setting, we observe an increment of +9.1 points with respect to SegFormer, with consistent gains at both the fine and coarse-level granularities. Interestingly, Mask2Former benefits even further from increasing the training crop size, gaining +5.9 points. This improvement is primarily driven by the improvement at the fine-grained level, which gains +7.4 points versus the +4.5 at the coarse level. These results further support our hypothesis that sufficiently large context windows are especially relevant for separating between similar classes that belong to the same coarse semantic categories. Moreover, the query-based Mask2Former segmentation approach appears particularly well suited to leverage the additional contextual information, as it allows the learned queries to better localize and classify coherent semantic regions. Lastly, performing TTA with three scaling factors ($\times0.75$, $\times1$, and $\times1.25$) and horizontal flipping yields an additional gain of +1.4 points, resulting in a final $mIoU_{comp}$ of 74.8\% in the validation set.

The performance of our final submission, based on Mask2Former trained with large crops and leveraging TTA during inference is reported in Table~\ref{tab:challenge_results}. According to the official Codabench evaluation, the proposed solution achieves an $mIoU_{comp}$ of 69.6\% on the test set, ranking 5th among the challenge final submissions. A closer look at the per-category results reveals that underrepresented categories such as \textit{Animal} and \textit{Water} remain particularly challenging, with $IoU$s of 26.9\% and 58.8\%, respectively. Conversely, more frequent categories such as \textit{Terrain}, \textit{Vegetation}, and \textit{Sky} all surpass 90\% $IoU$. Another interesting observation is the significant performance gap between categories such as \textit{Vehicle} (92.6\% $IoU$) and \textit{Object} (62.7\% $IoU$). Although both categories are sufficiently represented in the dataset, vehicles exhibit more distinct and consistent visual appearance, while \textit{Object} is a more heterogeneous categorization that encompasses widely different elements. As a result, the boundary between \textit{Object} and other categories is ambiguous, resulting in increased confusion and lower segmentation performance.

{
\setlength{\tabcolsep}{1.8pt}

\begin{table}[t]
\centering
\caption{Results for the final Mask2Former submission on the GOOSE 2D Fine-Grained Semantic Segmentation Challenge test set.}
\label{tab:challenge_results}

\begin{tabular}{ccc|ccccccccccc}
\toprule
\multicolumn{3}{c|}{$mIoU$ (\%)} &
\multicolumn{11}{c}{Per-Category $IoU$ (\%)} \\
\cmidrule(lr){1-3}
\cmidrule(lr){4-14}

\rotatebox{90}{Comp.} &
\rotatebox{90}{Coarse} &
\rotatebox{90}{Fine} &
\rotatebox{90}{Animal} &
\rotatebox{90}{Constr.} &
\rotatebox{90}{Human} &
\rotatebox{90}{Object} &
\rotatebox{90}{Road} &
\rotatebox{90}{Sign} &
\rotatebox{90}{Sky} &
\rotatebox{90}{Terrain} &
\rotatebox{90}{Veget.} &
\rotatebox{90}{Vehicle} &
\rotatebox{90}{Water} \\
\midrule

69.6 &
76.6 &
62.7 &
26.9 &
84.4 &
85.8 &
62.7 &
71.5 &
76.5 &
97.8 &
91.1 &
94.2 &
92.6 &
58.8 \\

\bottomrule
\end{tabular}
\end{table}
}

\subsection{Qualitative Analysis}
\begin{figure}[t]
    \centering
    \includegraphics[width=.98\linewidth]{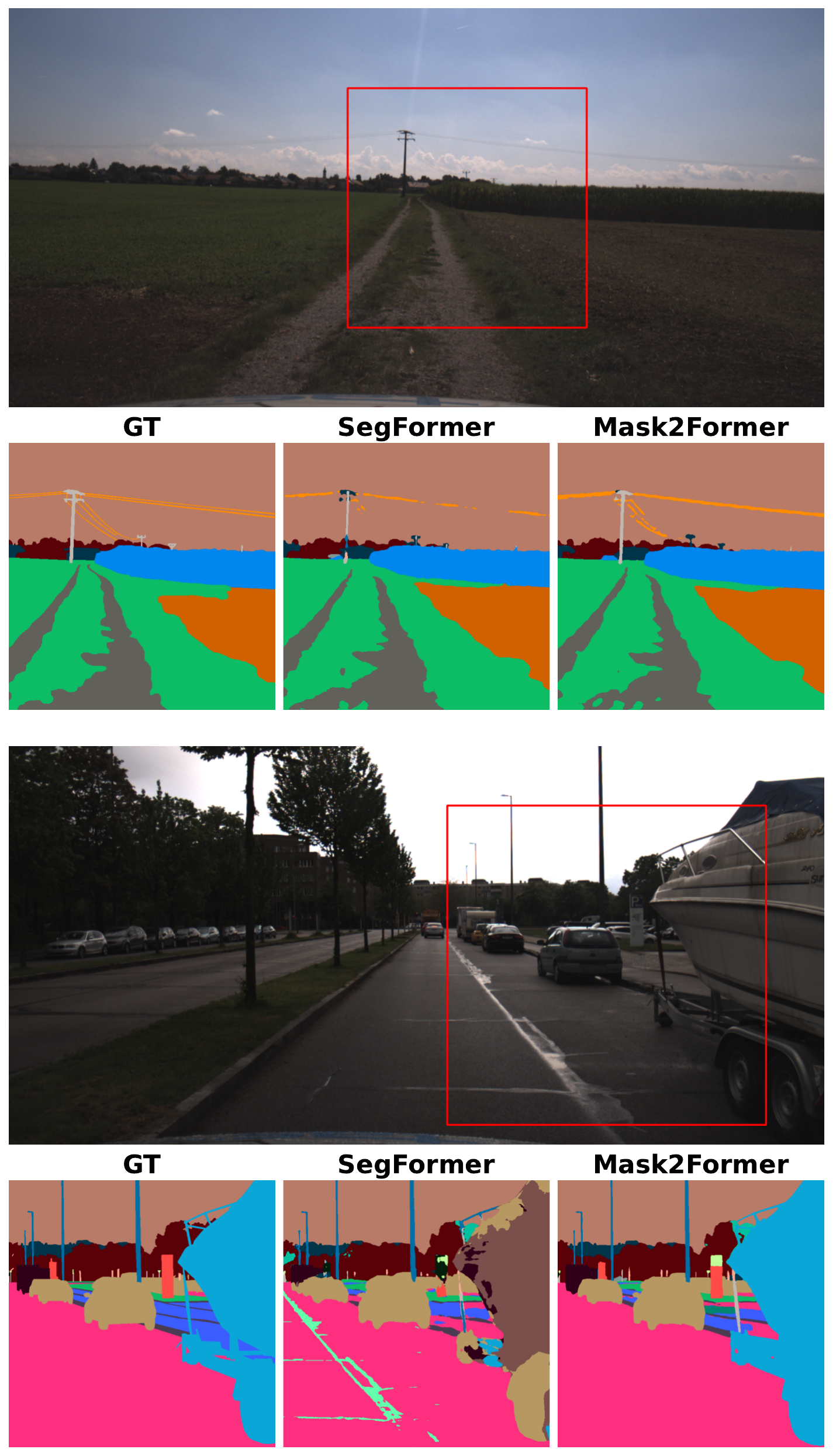} 
    \caption{Qualitative comparison between SegFormer and Mask2Former using the large-crop training configuration evaluated without TTA. Mask2Former is able to preserve thinner structures while yielding more spatially coherent masks.}
    \label{fig:model_comparison}
\end{figure}

In Figure~\ref{fig:model_comparison}, a qualitative comparison between SegFormer and Mask2Former trained with large crops and evaluated without TTA is shown. In the first sample, we observe that Mask2Former is able to preserve much thinner structures than SegFormer. This is particularly notable in the pole (\textit{gray}) and wires (\textit{light orange}) in the image, whose fine details and structure are better preserved by Mask2Former. We attribute this behavior to the pixel decoder in Mask2Former, which explicitly avoids the loss of high resolution features throughout the segmentation process.

Meanwhile, the second example highlights Mask2Former's ability to segment more coherent semantic regions. As seen in the trailer (\textit{light blue}), SegFormer not only misclassifies the object but yields noisy and fragmented masks that fail to capture its shape. In contrast, Mask2Former produces a coherent mask that accurately matches the object shape and boundaries. In the same sample, SegFormer's confusion of small wet regions on the road surface (\textit{magenta}) with lane markings (\textit{light green}) is revealing. Even though visually similar, the shape and spatial arrangement of such puddles are inconsistent with typical lane markings, a cue that Mask2Former is able to leverage to discard segmenting them as such.

\begin{figure}[t]
    \centering
    \includegraphics[width=.98\linewidth]{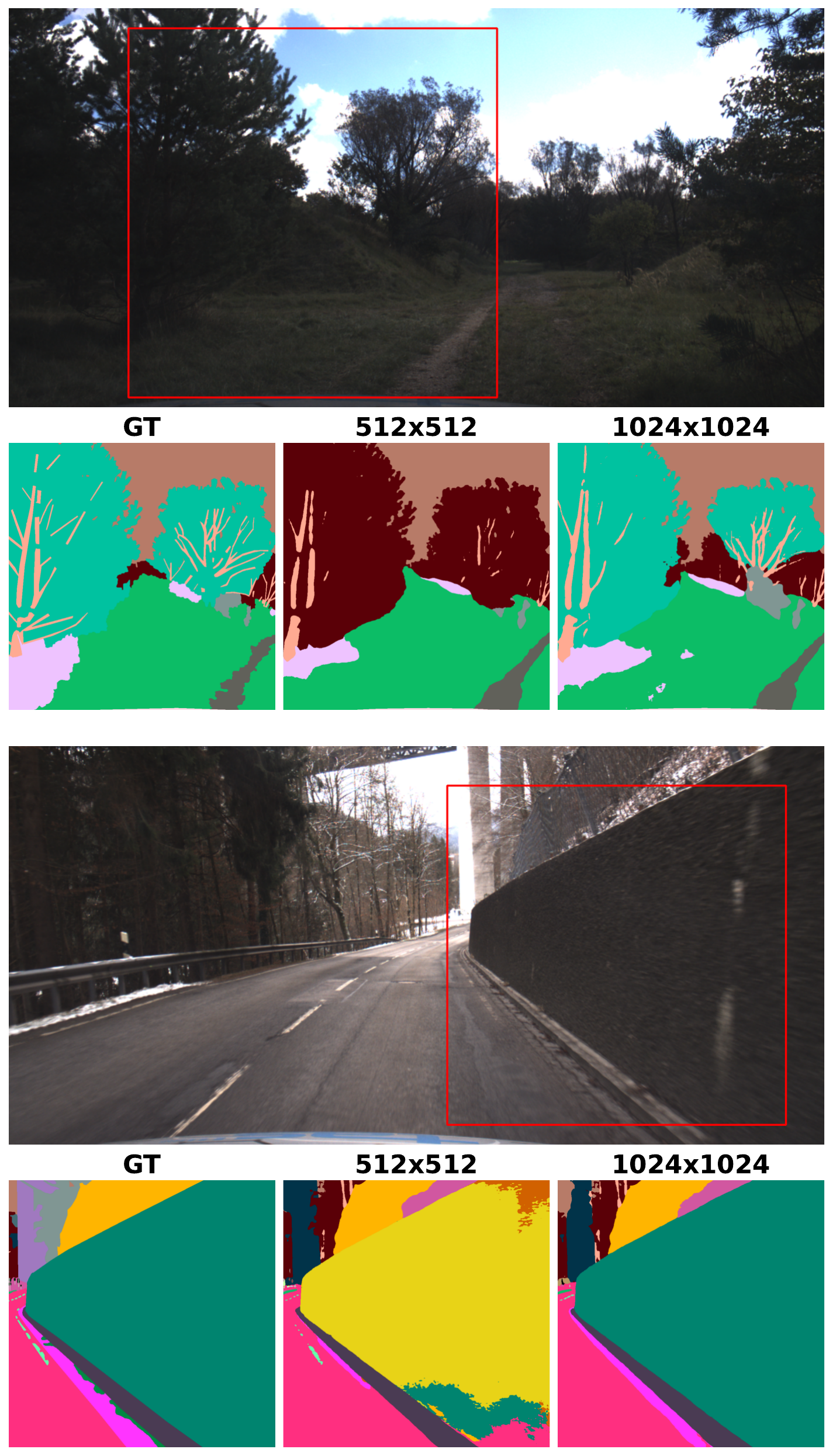} 
    \caption{Qualitative comparison between Mask2Former models trained with small ($512\times512$) and large ($1024\times1024$) crops and evaluated without TTA. Large-crop training allows the model to leverage scene-level context to accurately segment elements with visually ambiguous local appearances.}
    \label{fig:crop_comparison}
\end{figure}

In Figure~\ref{fig:crop_comparison}, we analyze the impact of increasing the spatial context available during training. To that end, we compare predictions from Mask2Former models trained with $512\times512$ and $1024\times1024$ crop sizes and no TTA. In the first example, an interesting behavior is observed. According to the GOOSE dataset labeling policy, a subtle distinction is made between the forest class (\textit{burgundy}), where individual tree structures are no longer discernible, and foreground trees explicitly annotated with their trunk (\textit{peach}) and crown (\textit{teal}) classes. While the model trained with smaller crops tends to confuse these visually similar categories, the model trained with the larger crop is able to reason over contextual cues such as the visibility of the tree trunk and the distance to the vehicle to discern between foreground trees and background forest.

Moreover, in the second sample, the wall (\textit{dark teal}) on the right side has an ambiguous appearance due to its dull texture and motion blur. Consequently, the model trained with the smaller crop misclassifies most of the structure as hedge (\textit{yellow}). In contrast, the model trained with the larger crop is able to accurately segment this structure. This observation further supports our hypothesis that increasing the spatial context available during training enables the model to leverage scene-level information, such as the presence of the road and the overall shape of the large wall structure, to provide a more reliable semantic interpretation.

\section{CONCLUSIONS}

In this technical report, we presented and analyzed our submission to the GOOSE 2D Fine-Grained Semantic Segmentation Challenge. Starting from a SegFormer baseline, we progressively improved our solution by transitioning to Mask2Former, increasing crop size during training, and, finally, performing TTA during inference.

Our results demonstrate that Mask2Former consistently outperforms SegFormer at both the coarse and fine-grained levels. Thanks to its query-based mask classification approach and its dual pixel and Transformer decoder architecture, Mask2Former is able to better preserve thin structures and produce more spatially coherent segmentation masks. Furthermore, increasing the crop size during training proves to have a substantial impact on segmentation quality. Access to broader spatial context enables the model to leverage scene-level cues and discern between visually similar elements that belong to different semantic categories. Lastly, TTA provides additional improvements by increasing the robustness of the final prediction.

Our final submission achieves an $mIoU_{comp}$ of 69.6\% on the challenge test set as reported by the Codabench evaluation platform, ranking 5th among all participating teams. Beyond providing a simple yet strong baseline for fine-grained semantic segmentation, our findings highlight the effectiveness of query-based mask classification models and demonstrate the critical role of spatial context in fine-grained scene understanding for outdoor environments.

\bibliography{bibliography}

\end{document}